\documentclass[conference]{IEEEtran}
\IEEEoverridecommandlockouts
\usepackage{cite}
\usepackage{amsmath,amssymb,amsfonts}
\usepackage{algorithm}
\usepackage{algpseudocode}
\usepackage{enumitem}

\usepackage{graphicx}
\usepackage{textcomp}
\usepackage{xcolor}

\usepackage{hyperref}
\usepackage{graphicx}  
\usepackage{amsthm}
\usepackage{booktabs}  
\usepackage{diagbox} 
\usepackage{tikz}
\usepackage{rotating}
\usepackage{bm}

\usepackage{float}

\algrenewcommand\algorithmicrequire{\textbf{Input:}}
\algrenewcommand\algorithmicensure{\textbf{Output:}}
\algnewcommand{\LeftComment}[1]{\Statex \(\triangleright\) #1}
\algrenewcommand{\algorithmiccomment}[1]{\hfill$\triangleright$ #1}

\theoremstyle{definition}
\newtheorem{problem}{\textsc{Problem}}
\newtheorem{definition}{\textsc{Definition}}

\theoremstyle{plain}
\newtheorem{lemma}{\textsc{Lemma}}

\theoremstyle{remark}

\usepackage[none]{hyphenat}
\usepackage{svg}

\usepackage{float}

\usepackage{hyperref}
\hypersetup{
    colorlinks=true,        
    linkcolor=gray,          
    citecolor=cyan,       
    citebordercolor = green,
    urlcolor=gray,           
}

\def\BibTeX{{\rm B\kern-.05em{\sc i\kern-.025em b}\kern-.08em
    T\kern-.1667em\lower.7ex\hbox{E}\kern-.125emX}}
    
\begin{document}

\title{Adaptive-GraphSketch: Real-Time Edge Anomaly Detection via Multi-Layer Tensor Sketching and Temporal Decay}

\IEEEaftertitletext{\vspace{-1.5\baselineskip}\noindent
\footnotesize This is the author’s version of the paper accepted to appear in the IEEE International Conference on Knowledge Graphs (ICKG 2025). 
The final published version will be available via IEEE Xplore.\par\vspace{1\baselineskip}}

\author{\IEEEauthorblockN{Ocheme Anthony Ekle}
\IEEEauthorblockA{\textit{Department of Computer Science} \\
\textit{Tennessee Technological University}\\
Cookeville, USA\\ 
oaekle42@tntech.edu
}
 \and
\IEEEauthorblockN{William Eberle}
\IEEEauthorblockA{\textit{Department of Computer Science} \\
\textit{Tennessee Technological University}\\
Cookeville, USA\\
weberle@tntech.edu}
}

\maketitle
\begin{abstract}
Anomaly detection in dynamic graphs is essential for identifying malicious activities, fraud, and unexpected behaviors in real-world systems such as cybersecurity and power grids. However, existing approaches struggle with scalability, probabilistic interpretability, and adaptability to evolving traffic patterns. In this paper, we propose \textsc{Adaptive-GraphSketch}, a lightweight and scalable framework for real-time anomaly detection in streaming edge data. Our method integrates temporal multi-tensor sketching with Count-Min Sketch using Conservative Update (CMS-CU) to compactly track edge frequency patterns with bounded memory, while mitigating hash collision issues. We incorporate Bayesian inference for probabilistic anomaly scoring and apply Exponentially Weighted Moving Average (EWMA) for adaptive thresholding tuned to burst intensity. Extensive experiments on four real-world intrusion detection datasets demonstrate that \textsc{Adaptive-GraphSketch} outperforms state-of-the-art baselines such as \textsc{AnoEdge-G/L}, \textsc{MIDAS-R}, and \textsc{F-FADE}, achieving up to 6.5\% AUC gain on CIC-IDS2018 and up to 15.6\% on CIC-DDoS2019, while processing 20 million edges in under 3.4 seconds using only 10 hash functions. Our results show that \textsc{Adaptive-GraphSketch} is practical and effective for fast, accurate anomaly detection in large-scale streaming graphs.
\end{abstract}


\section{Introduction}

Dynamic graph data is increasingly prevalent in real-time systems such as cybersecurity, social media, power grids, and fraud detection \cite{040_bhatia2022_MIDAS_latest, 105_shin2017_DenseAlert, ekle2024_dynamicGraph2024}. These systems generate massive, high-velocity streams of edges, representing relationships between nodes, and often exhibit evolving topologies and complex structural patterns. Detecting anomalies in such settings is critical for identifying malicious behavior, data breaches, and abnormal user activity.

However, traditional graph-based anomaly detection techniques, most based on personalized randomized walks \cite{AdaptiveDecayRank_ekle2025adaptive}, matrix factorization \cite{039c1_xie2023_multi_view}, and subgraph snapshot aggregation \cite{043_F-FADE_chang2021f,105_shin2017_DenseAlert}, struggle in streaming environments. They often rely on storing the full adjacency matrix or computing expensive subgraph statistics, leading to high memory overhead and delayed detection. Moreover, many existing models lack adaptability to rapid changes in network behavior and fail to provide interpretable, probabilistic outputs \cite{ekle2024_dynamicGraph2024}.

In this paper, we propose a lightweight, streaming anomaly detection framework called \textsc{Adaptive-GraphSketch}, which operates in real time without storing the full graph. Our method integrates temporal \textit{multi-tensor sketching}\cite{pham2013fast_tensorSketch} with Count-Min Sketch using Conservative Update (CMS-CU)\cite{estan2003new_CMS_Sketch_conservative_update} to compactly track edge frequencies using bounded memory, while mitigating hash collisions common in streaming settings. To enhance the detection of fast-changing anomalies, we incorporate Bayesian posterior scoring for uncertainty-aware inference and apply Exponentially Weighted Moving Average (EWMA) smoothing~\cite{lucas1990exponentially_moving_avg_EWMA} for dynamic thresholding. The EWMA parameters are adaptively tuned based on burst intensity (i.e., the rate of edge activity spikes within short time windows), enabling the model to adapt to concept drift and volatility in graph streams.

Unlike existing sketch-based models such as MIDAS-R \cite{040_bhatia2022_MIDAS_latest} and F-FADE \cite{043_F-FADE_chang2021f}, which lack probabilistic reasoning and struggle under bursty or rapidly evolving network dynamics, \textsc{Adaptive-GraphSketch} integrates temporal multi-tensor sketching, CMS-CU, Bayesian inference, and EWMA-based dynamic thresholding into a unified, real-time detection pipeline. To the best of our knowledge, it is the first edge-level streaming anomaly detection framework to combine these components into a probabilistic model with mathematically justified threshold adaptation. 

The key contributions of this work are:
\begin{itemize}
\item We propose a real-time anomaly detection framework that leverages temporal multi-tensor sketching to compactly track edge frequencies with bounded memory.
\item We integrate Count-Min Sketch with Conservative Update (CMS-CU) to efficiently track edge frequencies while mitigating hash collisions within the sketch tensor.
\item We introduce a Bayesian inference  for computing posterior anomaly scores that reflect    uncertainty and adaptivity to evolving graph behaviors.
\item We design an EWMA-based adaptive thresholding mechanism with burst-tuned smoothing to enhance robustness in volatile graph streams.
\item We conduct extensive experiments on four large-scale datasets.
\textsc{Adaptive-GraphSketch} delivers competitive performance on DARPA and ISCX-IDS2012 and achieves up to 6.5\% and 15.6\% AUC improvements on CIC-IDS2018 and CIC-DDoS2019 respectively, while processing 20 million edges in under 3.4 seconds with only 10 hash functions (i.e., row depth in the CMS-CU), demonstrating strong detection accuracy and runtime efficiency.
\end{itemize} 

The rest of this paper is organized as follows: Section~\ref{sec2:related_works} reviews related work; Section~\ref{sec3:problem_formation} presents the preliminaries and problem definition; Section~\ref{sec4:method} describes the methodology; Section~\ref{sec5:experiment} reports the experimental results; and Section~\ref{sec6:conclusion} concludes the paper with future directions.



\section{Related Work}\label{sec2:related_works}
\newcommand{\diagonalHeader}{
  \begin{tikzpicture}[scale=1]
    \draw[thick] (-0.1,2.1) -- (3.0,0); 
    \node[anchor=south west, font=\normalsize] at (0.1,0.1) {\text{Property}};
    \node[anchor=north east, font=\normalsize] at (2.9, 1.8) {\text{Method}};
  \end{tikzpicture}
}

  


\begin{table}[htbp]
  \centering
  \caption{\textbf{Our Method vs. baselines:} Comparison of \textsc{GraphSketch} with prior dynamic graph anomaly detection methods.}
  \label{tab1:relatedWork}
  \resizebox{\columnwidth}{!}{  
  \begin{tabular}{c|ccccccc}
    \toprule
    \diagonalHeader & 
    \rotatebox{90}{SedanSpot~\cite{101_eswaran2018_SedanSpot}} & 
    \rotatebox{90}{DenseStream~\cite{105_shin2017_DenseAlert}} & 
    \rotatebox{90}{PENminer~\cite{0103_belth2020_PENminer}} &
    \rotatebox{90}{F-FADE~\cite{043_F-FADE_chang2021f}} &
    \rotatebox{90}{MIDAS-R~\cite{040_bhatia2022_MIDAS_latest}} & 
    \rotatebox{90}{AnoEdge~\cite{056_bhatia2023_AnoEdge_sketch_based}} & 
    \rotatebox{90}{\textsc{Our Method}} \\
    \midrule
    Real-time detection$^*$     &  &  &  &  & \checkmark & \checkmark & \checkmark \\
    Edge anomalies              & \checkmark & \checkmark & \checkmark & \checkmark & \checkmark & \checkmark & \checkmark \\
    Temporal Tensor Sketching  &  &  &  &  &  &  & \checkmark \\
    Adaptive Bayesian Scoring  &  &  &  &  &  &  & \checkmark \\
    Uncertainty modeling       &  &  &  &  &  &  & \checkmark \\
    Memory-efficient           &  &  &  &  & \checkmark & \checkmark & \checkmark \\
    Sudden edge changes        & \checkmark &  & \checkmark & \checkmark & \checkmark & \checkmark & \checkmark \\
     \bottomrule
    \multicolumn{8}{l}{\footnotesize{$^*$Real-time = processing 20M edges within 10 seconds.}}
  \end{tabular}
  }
\end{table}

In this section, we review existing methods for detecting anomalies in dynamic graphs. We group prior work into three main categories based on their algorithmic strategies. Each class offers different trade-offs in terms of scalability, adaptability, and memory efficiency. For broader coverage, we refer readers to~\cite{028_sp_ranshous2015anomaly_detection_dynamic, ekle2024_dynamicGraph2024}.

\textbf{\textit{Edge Stream Methods}:} \textsc{SedanSpot}~\cite{101_eswaran2018_SedanSpot} detects sparse edge anomalies based on occurrence patterns. \textsc{PENminer}~\cite{0103_belth2020_PENminer} captures persistence in edge updates, while \textsc{F-FADE}~\cite{043_F-FADE_chang2021f} models frequency patterns via likelihood estimation. \textsc{MIDAS-R}~\cite{040_bhatia2022_MIDAS_latest} uses Count-Min Sketch with chi-squared testing for anomaly scoring. However, these approaches lack probabilistic reasoning, struggle with bursty behavior, and often require high computational cost. Our method addresses these gaps through multi-tensor sketching, Bayesian inference, and adaptive thresholding.

\textbf{\textit{Probabilistic Sketch Methods}:} Count-Min Sketch (CMS)~\cite{estan2003new_CMS_Sketch_conservative_update} has been used for scalable frequency estimation in graph streams. \textsc{RHSS}~\cite{111_ranshous2016scalable_CMS-Sketch} applies CMS to edge properties, while \textsc{AnoEdge}~\cite{056_bhatia2023_AnoEdge_sketch_based} uses higher-order sketching for count-based deviations. \textsc{DecayRank}~\cite{ekle2024dynamic_DecayRank} extends PageRank with temporal decay for node anomaly detection, and \textsc{Adaptive-DecayRank}~\cite{AdaptiveDecayRank_ekle2025adaptive} enhances this by using Bayesian updates and dynamic thresholds. While memory-efficient, these methods~\cite{040_bhatia2022_MIDAS_latest,056_bhatia2023_AnoEdge_sketch_based,111_ranshous2016scalable_CMS-Sketch,ekle2024dynamic_DecayRank} often rely on fixed thresholds and lack general uncertainty modeling, limiting performance in high-frequency edge streams.  In contrast, \textsc{Adaptive-GraphSketch} integrates 3D tensor sketching with conservative updates, Bayesian scoring, and dynamic EWMA-based thresholding for robust, real-time adaptation.

\textbf{\textit{Matrix Factorization Methods}:} \textsc{DenseStream}~\cite{105_shin2017_DenseAlert} incrementally tracks dense subtensors. \textsc{EdgeMonitor}~\cite{039b_wang2017_Edge-Monitoring} models edge transitions using first-order Markov chains. \textsc{MultiLAD}~\cite{039c1_xie2023_multi_view} uses spectral decomposition for subgraph anomalies. While effective offline, these methods are computationally expensive and unsuitable for real-time settings. In contrast, our method avoids global recomputation and uses bounded-memory summaries for efficient real-time edge anomaly detection.

As summarized in Table~\ref{tab1:relatedWork}, our method is the first to unify temporal sketching, conservative updates, Bayesian scoring, and adaptive thresholding in a single real-time pipeline. It offers a practical balance between scalability, interpretability, and resilience under volatile graph streams.

\section{Preliminaries and Problem Definition}
\label{sec3:problem_formation}
Let $E = \{e_1, e_2, \ldots\}$ be a stream of edges from a dynamic graph $G = (V, E)$, where each edge $e_i = (u_i, v_i, t_i)$ denotes an interaction from node $u_i$ to $v_i$ at time $t_i$. 
All notations used throughout this section and the rest of the paper are summarized in Table~\ref{tab2:notation}.

\begin{table}[htbp]
\caption{Summary of Notations Used in Adaptive-GraphSketch}
\label{tab2:notation}
\centering
\renewcommand{\arraystretch}{1.2}
\begin{tabular}{p{1.6cm} p{6cm}}
\toprule
\textbf{Symbol} & \textbf{Definition} \\
\midrule
$e = (u,v,t)$ & Edge from node $u$ to $v$ arriving at time $t$. \\
$a$ & Observed frequency of edge $(u,v)$ at time $t$. \\
$s$ & Cumulative frequency of edge $(u,v)$ before $t$. \\
$\hat{f}(x)$ & Estimated frequency of item $x$ via CMS-CU. \\
$d$ & Number of hash functions (rows in sketch). \\
$h_i(x)$ & Hash function $i$ applied to $x$. \\
$v_i$ & Counter value in row $i$ of the sketch. \\
$C_{i,j}^{(t)}$ & Value in sketch at row $i$, column $j$, time $t$.\\
$\boldsymbol{\mathcal{S}}_u$, $\boldsymbol{\mathcal{S}}_v$ & Hash-based sketches of nodes $u$ and $v$.
\\
\midrule
$\mu = \frac{s}{t}$ & Expected frequency under normal behavior. \\
$\sigma^2 = \frac{s}{t^2}$ & Variance estimate under normal behavior. \\
$\mu_A = \mu + \delta$ & Mean under anomaly assumption. \\
$\sigma_A^2 = 4\sigma^2$ & Increased variance under anomaly. \\
\midrule
$\lambda$ & Weight for smoothing past and current stats. \\
$\tilde{s}_t$ & Smoothed cumulative frequency at time $t$. \\
$\gamma$ & Decay factor ($0 < \gamma < 1$). \\
\midrule
$X_t$ & Anomaly score or posterior at time $t$. \\
$\alpha$ & EWMA smoothing constant ($0 < \alpha < 1$). \\
$\tau_t$ & Dynamic threshold at time $t$. \\
$\mu_t$ & Mean of $X_t$ scores up to $t$. \\
$\sigma_t$ & Std. deviation of $X_t$ scores up to $t$. \\
\bottomrule
\end{tabular}
\end{table}

\begin{figure}[htbp]
  \centering
  \includegraphics[width=0.48\textwidth, height=0.15\textheight]{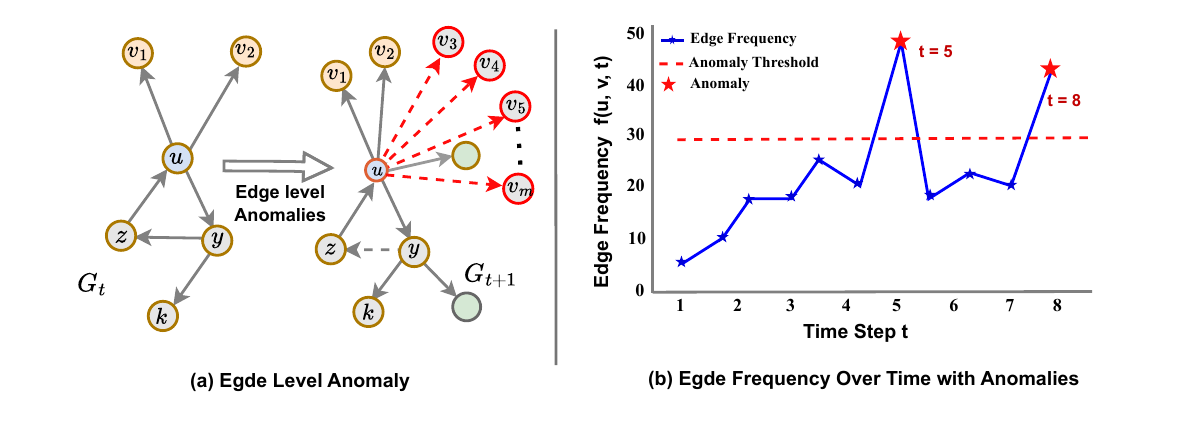}
  \caption{
  \textbf{Edge-level anomalies across two time steps} (\(t\) and \(t+1\)): 
  \textbf{(a)} illustrates sudden bursts in edge activity and temporal microcluster formations as the graph evolves from \(G_t\) to \(G_{t+1}\); 
  \textbf{(b)} shows rare and bursty edge frequency patterns that significantly deviate from historical trends, with spikes at \(t = 5\) and \(t = 8\) suggesting potential cyber attack events.
  }
  \label{fig:edge_anomalies}
\end{figure}

\begin{definition}
\textsc{(Edge-level Anomaly):} An edge-level anomaly occurs at time $t$ when the observed frequency of an edge $(u,v)$ significantly deviates from its historical average. Given time window $t \geq 1$ and threshold $\alpha > 0$, the anomaly score is:
\end{definition}
\begin{equation}
\label{eqn1:anomalyScore_edge_struture}
\text{Anomaly Score}(e, t) = \frac{(a - \frac{s}{t})^2 \cdot t}{s \cdot (t - 1)} \geq \alpha,
\end{equation}
where $a$ is the observed frequency at time $t$, $s$ is its cumulative historical frequency over previous windows, and $\alpha$ is the anomaly detection threshold.

As shown in Figure~\ref{fig:edge_anomalies}, edge-level anomalies may manifest as sudden bursts or emerging microclusters. For example, node $u$ rapidly connects to nodes $\{v_1, v_2, \dots, v_m\}$ at $G_{t+1}$, while rare edge spikes at $t=5$ and $t=8$ indicate unexpected frequency surges.

\subsection{Count-Min Sketch (CMS)}
CMS~\cite{111_ranshous2016scalable_CMS-Sketch} is a probabilistic data structure for frequency estimation using a 2D array of counters and $d$ hash functions. For an item $x$, its frequency estimate is computed as:
\begin{equation}
\hat{f}(x) = \min_{i=1}^{d} \text{CM}_{i,h_i(x)}.
\end{equation}
where $d$ is the number of hash functions, and $h_i(x)$ is the $i$-th hash function. CMS achieves significant memory reduction and offers sublinear memory usage but suffers from overestimate counts due to \textit{hash collisions} counter~\cite{mazziane2022_CMS_update_formal_analysis}, where multiple elements are mapped to the same.

\subsection{CMS with Conservative Update (CMS-CU)}
CMS-CU~\cite{estan2003new_CMS_Sketch_conservative_update} improved upon the standard CMS \cite{111_ranshous2016scalable_CMS-Sketch} by reducing overestimation errors from hash collisions. Unlike CMS, which increments all counters corresponding to hash functions,  CMS-CU only updates those counters with the current minimum value. Given an edge $e = (u, v)$, we maintain two sketches: $\text{CMS-CU}_{\text{current}}$ for the current time window and $\text{CMS-CU}_{\text{total}}$ for cumulative history.

Let $v_i$ be the value of the counter indexed by $h_i(x)$, we update only the counters equal to the row minimum:
\begin{equation}
v_i \leftarrow v_i + 1 \quad \text{if} \quad v_i = \min_{j=1}^{d} v_j,
\end{equation}

The frequency estimate of an element $x$ is obtained by retrieving the minimum counter value across all hash functions:
\begin{equation}
\hat{f}(x) = \min_{i=1}^{d} \text{CMS-CU}_{i,h_i(x)}.
\end{equation}

\subsection{Tensor Sketching}
\label{subsec:tensorSketch_optional}
Tensor Sketching~\cite{pham2013fast_tensorSketch} generalizes pairwise edge tracking to higher-order interactions. Given an edge $(u,v,t)$, its outer product is approximated in polynomial kernel space via sketch composition:
\begin{equation}
\hat{f}_{uv}^{(k)}(t) = \text{FFT}^{-1} \left( \text{FFT}(\boldsymbol{\mathcal{S}}_u) \circ \text{FFT}(\boldsymbol{\mathcal{S}}_v) \right),
\end{equation}
where $\boldsymbol{\mathcal{S}}_u$, $\boldsymbol{\mathcal{S}}_v$ are hash-based sketches of $u$ and $v$, and $\circ$ denotes element-wise multiplication. $\hat{f}_{uv}^{(k)}(t)$ captures $k$-order edge patterns in compressed form. While classical tensor sketching uses Fast Fourier Transform (FFT) for convolution, our method skips FFT by directly updating a 3D Count-Min Sketch, enabling real-time, memory-efficient edge tracking.

\subsection{Bayesian Anomaly Scoring}
Bayesian inference~\cite{m5_Bayesian_inference_wasserman2013all} offers a probabilistic framework for combining prior beliefs with observed data to estimate event likelihood. In anomaly detection, it enables the adaptive computation of the posterior probability that an observation (e.g., edge or node interaction) is anomalous based on past behavior. We define Bayes’ Theorem as:
\begin{equation}
P(\text{Anomaly} \mid a, \mathcal{H}) = \frac{P(a \mid \text{Anomaly}) \cdot P(\text{Anomaly})}{P(a)},
\end{equation}
where $a$ is the observed frequency and $\mathcal{H}$ is historical data. This allows adaptive, uncertainty-aware scoring.

\subsection{Problem Statement}
We aim to detect anomalies in a dynamic edge stream $E = \{e_1, e_2, \ldots\}$, where each edge $e_i = (u_i, v_i, t_i)$ represents an interaction from source $u_i \in V$ to destination $v_i \in V$ at time $t_i$. The goal is to assign an anomaly score to each edge based on its deviation from historical patterns.

\begin{problem}
\label{problem_1}
\textit{Given a streaming graph $E$, compute an anomaly score for each incoming edge $e = (u,v,t)$ by comparing its observed frequency with historical behavior. Higher scores reflect unusual edge behavior, such as rare spikes or rapid bursts of interactions.}
\end{problem}

An edge $e = (u,v,t)$ is flagged as \textit{anomalous} if its score exceeds a dynamic threshold $\alpha$, which adapts over time to balance sensitivity and false positive control, as shown in Figure~\ref{fig:edge_anomalies}.

\section{Method}
\label{sec4:method}
\begin{figure*}[htbp]
  \centering
  \includegraphics[width=0.95\textwidth, height=0.40\textheight]{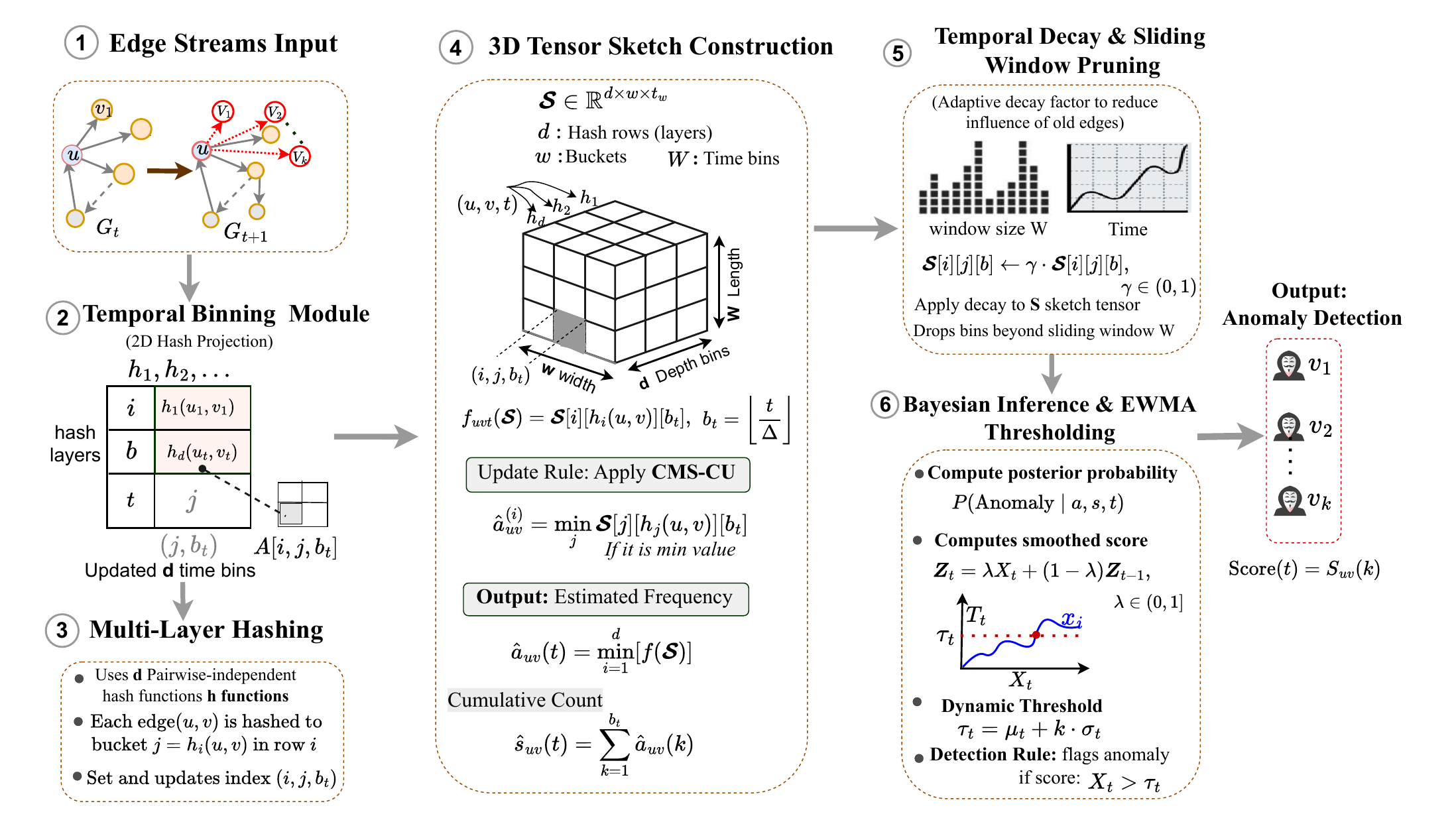}
  \caption{
  \textbf{The Framework of \textsc{Adaptive-GraphSketch.}}
  The model consists of six key phases: 
  (1) \textbf{Edge Streams Input:} Incoming edge events $(u,v,t)$ are observed as a dynamic graph stream. 
  (2) \textbf{Temporal Binning Module:} Edges are discretized into time bins $b_t = \lfloor t / \Delta \rfloor$ using hash-based projection. 
  (3) \textbf{Multi-Layer Hashing:} Each edge is projected into multiple hash layers via $j = h_i(u,v)$, producing $(i,j,b_t)$ indices. 
  (4) \textbf{3D Tensor Sketch Construction:} A sketch tensor $\mathcal{S} \in \mathbb{R}^{d \times w \times W}$ is updated using CMS-CU, enabling compact frequency tracking. 
  (5) \textbf{Temporal Decay and Sliding Window Pruning:} Applies decay factor $\gamma$ and maintains only the most recent $W$ time bins. 
  (6) \textbf{Bayesian Inference and EWMA Thresholding:} Uses posterior estimation and a dynamic threshold $\tau_t$ to flag anomalies with $\text{Score}_{uvt} > \tau_t$. 
  The output is a ranked anomaly list for edges in the stream.
  }
  \label{fig:framework}
\end{figure*}

In this section, we present \textsc{Adaptive-GraphSketch}, a lightweight, edge-level, real-time anomaly detection algorithm using a Multi-layer Tensor Sketch structure. The method accurately tracks temporal patterns in edge streams while maintaining low memory and computation overhead. The overall framework is illustrated in Figure~\ref{fig:framework}.

The key innovations in our method include:

\begin{itemize}
    \item \textbf{Multi-layer Tensor Sketching:} A compact 3D sketch $\boldsymbol{\mathcal{S}} \in \mathbb{R}^{d \times w \times t_w}$ encodes edge frequencies over time using hash-based layers.
    
    \item \textbf{Lightweight Frequency Estimation:} A 3D CMS with Conservative Updates (CMSCU) estimates counts across $(d,w,t_w)$ while reducing overestimation.
    
    \item \textbf{Temporal Decay and Pruning:} Applies exponential decay ($\gamma$) and sliding window to focus on recent activity.
    
    \item \textbf{Adaptive Bayesian Scoring:} Computes posterior score $P(\text{Anomaly} \mid a, s, t)$ from sketch-derived statistics.
    
    \item \textbf{Dynamic Thresholding:} Uses EWMA and $\tau_t = \mu_t + k\sigma_t$ for adaptive detection with smoothed thresholds.
\end{itemize}

\begin{figure}[htbp]
  \centering
  \includegraphics[width=0.48\textwidth, height=0.25\textheight]{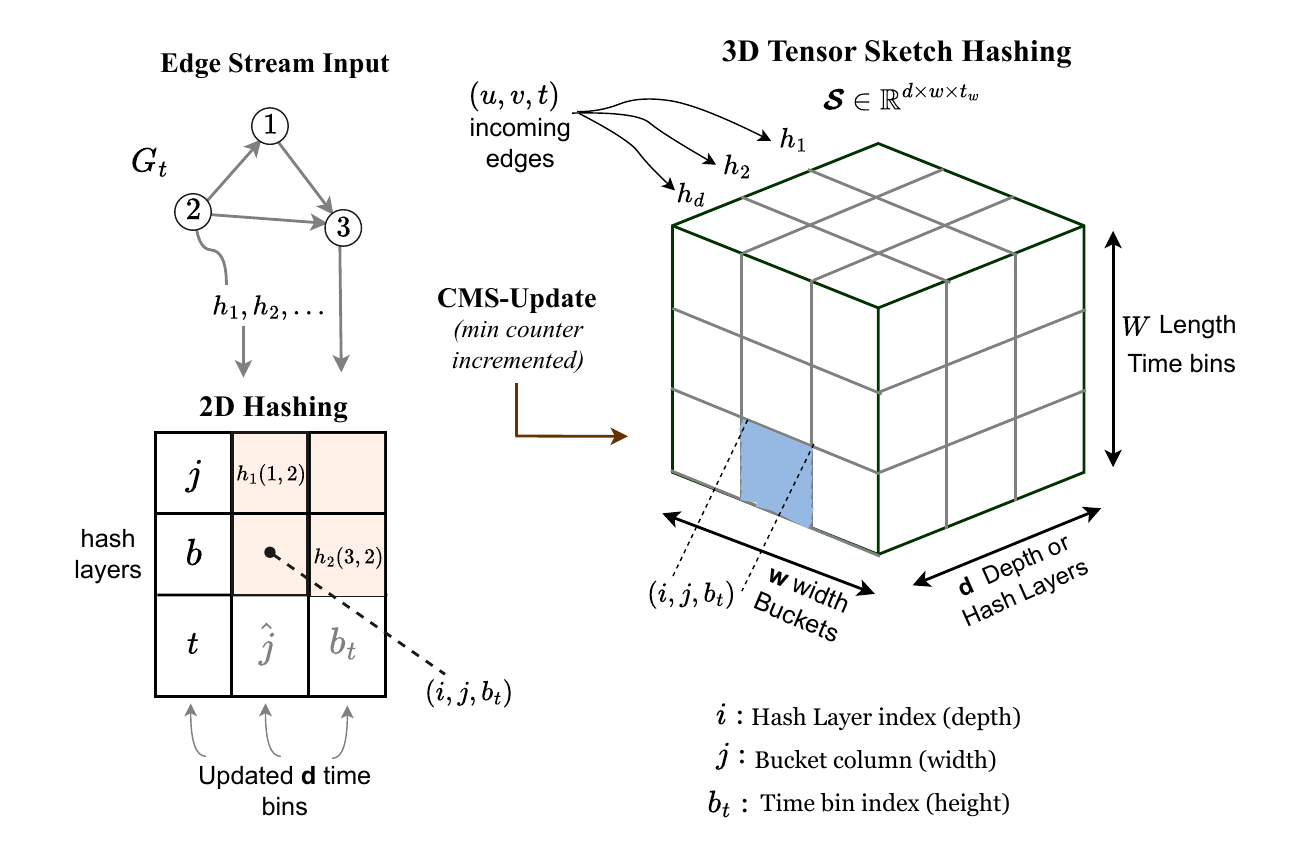}
  \caption{\textbf{Multi-Layer Tensor Sketch Overview}. An incoming edge $(u, v, t)$ is processed using multiple hash functions $h_1, h_2, \dots, h_d$, where each $h_i$ maps the edge to a bucket $j$ in the $i$-th sketch layer. The time dimension is discretized into bins via $b_t = \lfloor t / \Delta \rfloor$. The 3D sketch tensor $\mathcal{S}[i][j][b_t]$ is then updated using Count-Min Sketch with Conservative Update (CMS-CU), where only the minimum counter across all hash layers is incremented.} \label{fig:Multi_tensor-sketch}
\end{figure}

\subsection{Multi-Layer Tensor Sketching}
\label{subsec:tensorSketching}
To approximate evolving edge behavior over time, we introduce a \textbf{3D} tensor sketch structure $\boldsymbol{\mathcal{S}} \in \mathbb{R}^{d \times w \times W}$ that compactly encodes edge frequency dynamics using \textit{hash-based projection and time-aware indexing}. Inspired by prior work on tensor sketching for efficient high-dimensional approximation via randomized hashing \cite{pham2013fast_tensorSketch}.  We introduce a third axis for time-aware indexing $t_w$, our tensor sketch captures edge dynamics across $d$ hash rows, $w$ buckets, and $W$ time bins with bounded memory. This process is illustrated in Figure~\ref{fig:Multi_tensor-sketch}, where the incoming edge $(u,v,t) \in \mathcal{E}$ is hashed across multiple layers and projected into a 3D sketch tensor indexed by time.

We define the dynamic graph stream as a sequence of timestamped edge events $(u,v,t) \in \mathcal{E}$. 
To model temporal evolution, we divide the timeline into fixed-size intervals of width $\Delta$, and assign each edge to a corresponding time bin:
\begin{equation}
b_t = \left\lfloor \frac{t}{\Delta} \right\rfloor, \quad b_t \in \{1, 2, \dots, W\},
\end{equation}
where, $t$ is the timestamp of edge $(u, v)$ in the stream, ${\Delta}$ is the time bin width, and $W$ the maximum number of active time bins maintained in the tensor sketch $\boldsymbol{\mathcal{S}} \in \mathbb{R}^{d \times w \times W}$.

We use a family of \(d\) pairwise-independent hash functions \(\{ h_1, h_2, \dots, h_d \}\), where each \(h_i: \mathbb{N} \times \mathbb{N} \to \{1, 2, \dots, w\}\) maps an edge \((u,v)\) to a \textit{column bucket} in the \(i\)-th hash row. The sketch cell \(\boldsymbol{\mathcal{S}}[i, h_i(u,v), b_t]\) stores the interaction frequency of edge \((u,v)\) in the corresponding time bin.

Each edge arrival $(u,v,t)$, triggers a \textit{hash-based projection} into all $d$ rows of the tensor sketch by updating:
\begin{equation}
\boldsymbol{\mathcal{S}}[i][h_i(u,v)][b_t] \leftarrow \boldsymbol{\mathcal{S}}[i][h_i(u,v)][b_t] + 1,
\end{equation}
for $i = 1$ to $d$.

We adopt the conservative update (CMS-CU) mechanism \cite{estan2003new_CMS_Sketch_conservative_update}, where only sketch counters with the current minimum value among $\{\boldsymbol{\mathcal{S}}[j][h_j(u,v)][b_t]\}_{j=1}^d$ are incremented. 
\begin{equation}
\text{If } \boldsymbol{\mathcal{S}}[i][h_i(u,v)][b_t] = \min_j \boldsymbol{\mathcal{S}}[j][h_j(u,v)][b_t],
\end{equation}
where $h_i(u,v)$ is the $i$-th hash function mapping edge $(u,v)$ to a sketch column, and $b_t = \left\lfloor \frac{t}{\Delta} \right\rfloor$ is the current time bin index for timestamp $t$.

The estimated frequency of edge $(u,v)$ in the current bin is defined as:
\begin{equation}
\hat{a}_{uv}(t) = \min_{i=1}^{d} \boldsymbol{\mathcal{S}}[i][h_i(u,v)][b_t],
\end{equation}
where $\hat{a}_{uv}(t)$ denotes the estimated occurrence count of edge $(u,v)$ in time bin $b_t$.

To maintain long-term context, we define the cumulative frequency of the edge as:
\begin{equation}
\hat{s}_{uv}(t) = \sum_{k=1}^{b_t} \hat{a}_{uv}(k),
\end{equation}
where $\hat{s}_{uv}(t)$ captures the aggregated frequency of edge $(u,v)$ over all bins up to $b_t$.

Unlike traditional \textbf{2D} adjacency matrices $\boldsymbol{X} \in \mathbb{R}^{d \times w}$, which lack temporal granularity and grow linearly with graph size, our \textbf{3D} tensor sketch $\boldsymbol{\mathcal{S}} \in \mathbb{R}^{d \times w \times W}$ maintains a fixed memory bound of $\mathcal{O}(d \cdot w \cdot W)$ and enables real-time edge tracking with \textit{sublinear space}. Edge updates are processed in constant time, $\mathcal{O}(1)$. By organizing the sketch along discrete time bins, our approach supports efficient \textit{temporal querying}, decay-based forgetting, and scalable sliding window maintenance.

Algorithm~\ref{alg:adaptive-graphsketch} summarizes the complete edge processing pipeline, including tensor updates, decay, sketching, and frequency estimation.

\subsection{Frequency Estimation with Conservative Updates}
\label{subsec:cmscu}

We employ CMS-CU \cite{estan2003new_CMS_Sketch_conservative_update} as a lightweight and memory-efficient data structure for tracking edge frequencies in graph streams. For a given edge $e = (u,v,t)$, the \textit{estimated frequency} of edge $(u,v)$ at time $t$ is denoted by $\hat{a}_{uv}(t)$, while the \textit{cumulative historical frequency} up to $t$ is $\hat{s}_{uv}(t)$.

To reduce bias from hash collisions, CMS-CU selectively increments only the counters with the current minimum values:
\begin{equation}
v_i \leftarrow v_i + 1 \quad \text{if} \quad v_i = \min_{j=1}^{d} v_j,
\end{equation}
The estimated frequency of item $x$ is then:
\begin{equation}
\hat{f}(x) = \min_{i=1}^{d} \text{CMS-CU}_{i,h_i(x)}.
\end{equation}

\subsection{Raw Anomaly Score Based on Frequency Deviation}
\label{subsec:rawScore}

Once the current frequency $\hat{a}_{uv}(t)$ and cumulative count $\hat{s}_{uv}(t)$ are estimated, we compute a \textit{Raw Anomaly Score} to quantify how much recent edge activity deviates from expected behavior.

Let $a = \hat{a}_{uv}(t)$ and $s = \hat{s}_{uv}(t)$. The raw anomaly score is:
\begin{equation}
\text{RawScore}(u,v,t) = \frac{(a - \frac{s}{t})^2 \cdot t}{s \cdot (t - 1)}.
\label{eqn:rawScore}
\end{equation}
This formulation captures the squared deviation of current activity from historical average, normalized by variance, and highlights bursts or drops in edge behavior.






\subsection{Adaptive Probabilistic Scoring via Bayesian Inference}
\label{subsec:bayesian}

To incorporate adaptivity and capture uncertainty, we extend the score in Equation~\ref{eqn:rawScore} into a  Bayesian framework, modeling short-term activity against long-term trends. We then assume the mean and variance under normal behavior:
\[
\mu = \frac{s}{t}, \quad \sigma^2 = \frac{s}{t^2}
\]

\begin{enumerate}[label=\arabic*.]
    \item \textit{Likelihood under Normal:}
    \begin{equation}
    P(a \mid \text{Normal}) = \frac{1}{\sqrt{2\pi \sigma^2}} \exp\left( -\frac{(a - \mu)^2}{2\sigma^2} \right)
    \end{equation}

    \item \textit{Likelihood under Anomaly:} 
    We model anomalous behavior by shifting the mean and inflating the variance:
    \[
    \mu_A = \mu + \delta, \quad \sigma_A^2 = 4\sigma^2
    \]
    \begin{equation}
    P(a \mid \text{Anomaly}) = \frac{1}{\sqrt{2\pi \sigma_A^2}} \exp\left( -\frac{(a - \mu_A)^2}{2\sigma_A^2} \right)
    \end{equation}

    
    \item \textit{Posterior Inference:}     Assuming a fixed prior $P(\text{Anomaly}) = p_0$, the posterior probability is computed via Bayes’ rule:
    \begin{equation}
    P(\text{Anomaly} \mid a) = \frac{P(a \mid \text{Anomaly}) \cdot p_0}{P(a)},
    \end{equation}
    where
    $ P(a) = P(a \mid \text{Anomaly}) \cdot p_0 + P(a \mid \text{Normal}) \cdot (1 - p_0),
    $
    and the posterior gives an adaptive and normalized anomaly score for edge $(u,v)$ at time $t$, reflecting both statistical deviation and uncertainty in observed behavior.
\end{enumerate}

\subsection{Streaming Temporal Decay and Pruning}
\label{subsec:phase3}

To emphasize recency and maintain bounded memory, we incorporate two temporal mechanisms into the sketch: (i) exponential decay and (ii) sliding window pruning. 


\begin{algorithm}[H]
\caption{\textsc{Adaptive-GraphSketch}: Multi-Layer Tensor Sketching and Edge Processing}
\label{alg:adaptive-graphsketch}
\begin{algorithmic}[1]
\Require stream of edges $\{(u, v, t)\}$ over time
\Ensure real-time anomaly score per edge $(u, v, t)$

\State \textbf{initialize:} 3D tensor sketch $\boldsymbol{\mathcal{S}} \in \mathbb{R}^{d \times w \times W}$ with zeros
\State Set decay factor $\gamma \in (0,1]$, bin size $\Delta$, window size $W$

\State \textbf{for each} edge $(u,v,t)$ in stream:

    \State $b_t \gets \left\lfloor \frac{t}{\Delta} \right\rfloor$ \Comment{\textit{time bin index}}

    \Statex \textit{// multi-layer cms-cu sketch update}
    \For{$i = 1$ to $d$}
        \State $j \gets h_i(u, v)$ \Comment{\textit{hash index}}
        \If{$\boldsymbol{\mathcal{S}}[i][j][b_t] = \min_j \boldsymbol{\mathcal{S}}[j][h_j(u, v)][b_t]$}
            \State $\boldsymbol{\mathcal{S}}[i][j][b_t] \gets \boldsymbol{\mathcal{S}}[i][j][b_t] + 1$
        \EndIf
    \EndFor

    \Statex \textit{// apply temporal decay}
    \For{$b = 1$ to $W$}
        \State $\boldsymbol{\mathcal{S}}[i][j][b] \gets \gamma \cdot \boldsymbol{\mathcal{S}}[i][j][b]$
        \If{$b_t - b > W$}
            \State $\boldsymbol{\mathcal{S}}[i][j][b] \gets 0$
        \EndIf
    \EndFor

    \Statex \textit{// estimate edge frequencies}
    \State $\hat{a}_{uv}(t) \gets \min_{i=1}^{d} \boldsymbol{\mathcal{S}}[i][h_i(u, v)][b_t]$
    \State $\hat{s}_{uv}(t) \gets \sum_{k=1}^{b_t} \hat{a}_{uv}(k)$

    \Statex \textit{// anomaly score computation}
    \State $\text{Score}(u, v, t) \gets \texttt{BayesianScore}(\hat{a}_{uv}(t), \hat{s}_{uv}(t), t)$
    \State \Return $\text{Score}(u, v, t)$
\end{algorithmic}
\end{algorithm}

\paragraph{Exponential Temporal Decay.}  
Before ingesting new edge updates at time $t$, the sketch tensor $\boldsymbol{\mathcal{S}} \in \mathbb{R}^{d \times w \times W}$ is decayed along the temporal axis to progressively diminish the influence of older interactions. This is achieved by scaling each counter by a decay factor $\gamma \in (0,1)$:

\begin{equation}
\boldsymbol{\mathcal{S}}[i][j][b] \leftarrow \gamma \cdot \boldsymbol{\mathcal{S}}[i][j][b], \quad \gamma \in (0,1),
\end{equation}
\[
\quad \forall~i \in [1,d],~j \in [1,w],~b \in [1,W],
\]
where $b$ indexes the time bin, and $\gamma$ controls the rate at which old edges fades. Smaller values of $\gamma$ cause faster decay, giving more weight to recent edge activity. This mechanism mimics a form of \textit{temporal forgetting} \cite{AdaptiveDecayRank_ekle2025adaptive}, where recent events dominate.

\paragraph{Sliding Window Pruning.}  
To further constrain memory usage, we adopt a fixed-size sliding window strategy that retains only the most recent $W$ time bins (i.e., snapshots of edge activity). Let $\bm{b_t}$ be the current time bin index at time $t$, and $b$ represent an existing bin in the sketch. Any bin that falls outside the window horizon is zeroed out:
\begin{equation}
\text{If } b_t - b > W \Rightarrow \boldsymbol{\mathcal{S}}[i][j][b] \leftarrow 0
\end{equation}
This pruning operation removes stale sketch slices entirely, freeing memory and removing obsolete edge history. It also prevents the anomaly scoring logic from being biased by long-term drift or noise. Together, decay and pruning act as a dual temporal filter. 





Algorithm~\ref{alg:bayesian-scoring} outlines how the posterior anomaly score is computed for each edge based on sketch-derived statistics.

\begin{algorithm}[H]
\caption{Bayesian Posterior Anomaly Scoring}
\label{alg:bayesian-scoring}
\begin{algorithmic}[1]
\Require Estimated frequency $\bm{a}$, cumulative $\bm{s}$, time bin $\bm{t}$
\Ensure Posterior anomaly score $P(\text{Anomaly} \mid a)$

\State Compute historical mean: $\mu \gets \frac{s}{t}$
\State Compute variance: $\sigma^2 \gets \frac{s}{t^2}$

\Statex \textit{// likelihood under normal behavior}
\State $P_{\text{normal}} \gets \frac{1}{\sqrt{2\pi \sigma^2}} \exp\left( -\frac{(a - \mu)^2}{2\sigma^2} \right)$

\Statex \textit{// likelihood under anomaly}
\State $\mu_A \gets \mu + \delta$
\State $\sigma_A^2 \gets 4\sigma^2$
\State $P_{\text{anomaly}} \gets \frac{1}{\sqrt{2\pi \sigma_A^2}} \exp\left( -\frac{(a - \mu_A)^2}{2\sigma_A^2} \right)$

\Statex \textit{// posterior computation}
\State $p_0 \gets 0.05$ \Comment{\textit{prior}}
\State $P(a) \gets p_0 \cdot P_{\text{anomaly}} + (1 - p_0) \cdot P_{\text{normal}}$
\State \Return $P(\text{Anomaly} \mid a) \gets \frac{p_0 \cdot P_{\text{anomaly}}}{P(a)}$
\end{algorithmic}
\end{algorithm}







\subsection{Dynamic Thresholding with EWMA and FPR Control}
\label{subsec:phase4}

Let \( X_t = \text{Score}(u,v,t) \) denote the anomaly score for edge \( (u,v) \) at time \( t \). To mitigate noisy fluctuations and control false positives in edge streams, we apply an adaptive thresholding mechanism based on the Exponentially Weighted Moving Average (EWMA) \cite{lucas1990exponentially_moving_avg_EWMA}. The smoothed anomaly signal \( \boldsymbol{Z}_t \) is recursively defined as:
\begin{equation}
\boldsymbol{Z}_t = \lambda X_t + (1 - \lambda) \boldsymbol{Z}_{t-1}, \quad \lambda \in (0,1],
\end{equation}

\noindent where $\lambda \in (0,1] $  controls the memory decay; larger values assign more weight to recent anomalies. The initial value is set as \( \boldsymbol{Z}_0 = X_1 \).

To adaptively estimate the decision boundary, we compute the empirical mean \( \mu_t \) and standard deviation \( \sigma_t \) of the past scores up to time \( t \), where:
\begin{equation}
\mu_t = \frac{1}{t} \sum_{i=1}^{t} X_i, \qquad 
\sigma_t^2 = \frac{1}{t} \sum_{i=1}^{t} (X_i - \mu_t)^2
\end{equation}

The dynamic threshold \( \boldsymbol{\tau_t} \) is then defined as:
\begin{equation}
{\tau_t} = \mu_t + k \cdot \sigma_t,
\end{equation}
\noindent where \( k \) is a sensitivity multiplier.

\paragraph{Detection Rule.} An edge is flagged as anomalous at time \( t \) if:
\begin{equation}
X_t > \tau_t,
\end{equation}
\noindent where \( X_t = \text{Score}(u,v,t) \) denotes the anomaly score for edge \( (u,v) \) at time \( t \).

\paragraph{False Positive Guarantee.} Using Chebyshev's inequality \cite{08b_mendenhall2012probability_chebyshev_ineq}, the probability of false alarm is bounded as:

\begin{equation}
\mathrm{FPR}_t \leq \frac{1}{k^2}
\label{eq:chebyshev_bound}
\end{equation}

The statistical guarantee in Equation~\ref{eq:chebyshev_bound}
 provides an upper bound on false positive rates, even under non-Gaussian noise. 
 
 Algorithm~\ref{alg:dynamic-thresholding} implements the adaptive thresholding procedure using EWMA smoothing and empirical variance for robust decision-making.

\begin{algorithm}[H]
\caption{Dynamic Thresholding and Anomaly Detection}
\label{alg:dynamic-thresholding}
\begin{algorithmic}[1]
\Require Anomaly score stream $\{X_t\}$, smoothing factor $\lambda \in (0,1]$, sensitivity $k$
\Ensure Anomaly decision flag per timestamp $t$

\State Initialize EWMA: $\boldsymbol{Z}_0 \gets X_1$ \Comment{Initial smoothed score}
\State Initialize $\mu_0 \gets X_1$, $\sigma_0 \gets 0$

\For{$t = 2$ to $T$}
    \State $\boldsymbol{Z}_t \gets \lambda \cdot X_t + (1 - \lambda) \cdot \boldsymbol{Z}_{t-1}$
    \State $\mu_t \gets \frac{1}{t} \sum_{i=1}^{t} X_i$
    \State $\sigma_t \gets \sqrt{\frac{1}{t} \sum_{i=1}^{t} (X_i - \mu_t)^2}$
    \State $\tau_t \gets \mu_t + k \cdot \sigma_t$
    \If{$\boldsymbol{Z}_t > \tau_t$}
        \State Flag edge $(u, v)$ at time $t$ as anomalous
    \Else
        \State Mark edge as normal
    \EndIf
\EndFor

\State \Return Anomaly flag per edge $(u, v, t)$
\end{algorithmic}
\end{algorithm}





\begin{lemma}[\textsc{False Positive Control via Adaptive Thresholding}]
\label{lemma:ewma-fp-control}
Let $\{X_t\}_{t=1}^T$ be the sequence of anomaly scores generated by a streaming detection model. Assume that the distribution of $X_t$ is unimodal, has finite variance, and is approximately stationary over short windows. Then, the adaptive thresholding rule
\[
\tau_t = \mu_t + k \cdot \sigma_t
\]
ensures that the false positive rate (FPR) at time $t$ is bounded by the tail probability of the underlying distribution:
\begin{equation}
\mathrm{FPR}_t \leq \Pr(X_t > \mu_t + k \cdot \sigma_t).
\end{equation}
\end{lemma}

\noindent\textbf{\textit{Proof}.}
We apply Chebyshev’s inequality \cite{08b_mendenhall2012probability_chebyshev_ineq} to the anomaly score \(X_t\), which has empirical mean \(\mu_t\) and standard deviation \(\sigma_t\) computed up to time \(t\). For any \(k > 0\), Chebyshev’s inequality guarantees:
\begin{equation}
\Pr(|X_t - \mu_t| \geq k \cdot \sigma_t) \leq \frac{1}{k^2}.
\label{eq:chebyshev_applied}
\end{equation}

Since the detection rule only flags anomalies when \(X_t > \tau_t = \mu_t + k \cdot \sigma_t\), the relevant tail probability is:
\begin{equation}
\Pr(X_t > \mu_t + k \cdot \sigma_t) \leq \frac{1}{k^2}.
\label{eq:fp_bound}
\end{equation}

Therefore, the likelihood of falsely classifying a normal edge as anomalous is bounded by:
\begin{equation}
\Pr(\text{False Positive}) \leq \frac{1}{k^2}.
\label{eq:fp_final}
\end{equation}

This result holds regardless of the exact shape of the distribution, provided it is unimodal and has finite variance. For instance, if \(k = 2\), then \(\mathrm{FPR}_t \leq 0.25\); and if \(k = 3\), then \(\mathrm{FPR}_t \leq 0.11\).
\hfill\qedsymbol


\subsection{Runtime and Output Tracking}
\label{subsec:runtime}

Finally, each detected anomaly is logged as:
\[
(u, v, t, \text{Score}_{uvt}) \quad \text{if } \text{Score}_{uvt} > \tau_t,
\]
where \(\text{Score}_{uvt}\) denotes the computed anomaly score for edge \((u,v)\) at time \(t\).

Let \( N = |\mathcal{E}| \) be the total number of edges processed in the stream. The average processing time per edge is computed as:

\begin{equation}
\text{AvgTime} = \frac{T_{\text{exec}}}{N},
\end{equation}

where \( T_{\text{exec}} \) is the total execution time for the entire stream.

This completes the pipeline of \textsc{AdaptiveGraphSketch}, enabling scalable and low-latency anomaly detection in high-velocity edge streams.

\section{Experiments}
\label{sec5:experiment}
We now evaluate \textsc{Adaptive-GraphSketch} for near real-time anomaly detection in dynamic graphs. The evaluation focuses on three key aspects: accuracy (ROC-AUC), runtime efficiency, and scalability across diverse graph structures.

\begin{table}[ht]
\centering
\caption{Dataset statistics.}
\label{tab:dataset_stats}
\begin{tabular}{lrrr}
\toprule
\textbf{Dataset} & \textbf{Nodes ($|V|$)} & \textbf{Edges ($|E|$)} & \textbf{Timestamps ($|T|$)} \\
\midrule
DARPA          & 25,525     & 4,554,344     & 46,567  \\
ISCX-IDS2012   & 30,917     & 1,097,070     & 165,043 \\
CIC-IDS2018    & 33,176     & 7,948,748     & 38,478  \\
CIC-DDoS2019   & 1,290      & 20,364,525    & 12,224  \\
\bottomrule
\end{tabular}
\end{table}

\begin{table*}[!htbp]
    \centering
    \caption{AUC and Running Time when Detecting Edge Anomalies (Average over 5 runs)}
    \label{tab:auc_runtime_table}
    \resizebox{\textwidth}{!}{
    \begin{tabular}{l|cc|cc|cc|cc}
    \toprule
    \textbf{Algorithm} & \textbf{DARPA} & \textbf{Time (s)} & \textbf{ISCX-IDS2012} & \textbf{Time (s)} & \textbf{CIC-IDS2018} & \textbf{Time (s)} & \textbf{CIC-DDoS2019} & \textbf{Time (s)} \\
    \midrule
    \textsc{DenseStream}       & 0.5323 $\pm{0.000}$ & 25.36    & 0.551 $\pm{0.000}$ & 92.54   & 0.756 $\pm{0.000}$ & 5186.9  & 0.263 & 99.78   \\
    \textsc{SedanSpot}         
                      & 0.6408 $\pm$ 0.0025 & 78.13  
                      & 0.5807 $\pm$ 0.0014 & 10.29  
                      & 0.3413 $\pm$ 0.0341 & 110.02  
                      & 0.5679 $\pm$ 0.0022 & 397.40 \\
    \textsc{MIDAS-R}           
                      & 0.9493 $\pm$ 0.0006 & 0.203  
                      & 0.7176 $\pm$ 0.0696 & 0.294  
                      & 0.8834 $\pm$ 0.0011 & 0.361  
                      & 0.9625 $\pm$ 0.0016 & 0.409  \\
    PENminer          & 0.872 & 18756   & 0.530 & 4680    & 0.821 & 36000  & ---   & $>$86400 \\
    \textsc{F-FADE}           
                      & 0.9173 $\pm$ 0.0041 & 132.29  
                      & 0.5100 $\pm$ 0.0165 & 42.36  
                      & 0.8432 $\pm$ 0.0038 & 98.04  
                      & 0.1499 $\pm$ 0.1178 & 30.3   \\
    \textsc{AnoEdge-G}         & \textbf{0.970} $\pm$ 0.001 & 21.47   & \textbf{0.954} $\pm$ 0.000 & 6.6329     & 0.975 $\pm$ 0.001 & 49.538   & 0.997 $\pm$ 0.001 & 92.85  \\
    \textsc{AnoEdge-L}         & \textbf{0.963} $\pm$ 0.001 & \textbf{0.277 }   & \textbf{0.950} $\pm$ 0.000 & 0.58     & 0.927 $\pm$ 0.035 & \textbf{0.4803}   & 0.998 $\pm$ 0.000 & \textbf{0.83}   \\
    \midrule
    \textbf{a\textsc{GraphSketch}} 
                      & \textbf{0.9568 $\pm$ 0.000} & \textbf{2.15 }
                      & \textbf{0.8761 $\pm$ 0.000}  & \textbf{0.5168 }
                      & \textbf{0.9923 $\pm$ 0.000} & \textbf{4.2006}
                      & \textbf{0.9993 $\pm$ 0.000}  & \textbf{8.948 } \\
    \bottomrule
    \multicolumn{9}{l}{\footnotesize *All experiments are repeated 5 times. We report the mean ± standard deviation of ROC-AUC and the mean runtime in seconds.}
    \end{tabular}
    }
\end{table*}

\begin{figure*}[!htbp]
    \centering
    \includegraphics[width=0.95\textwidth, height=0.48\textheight]{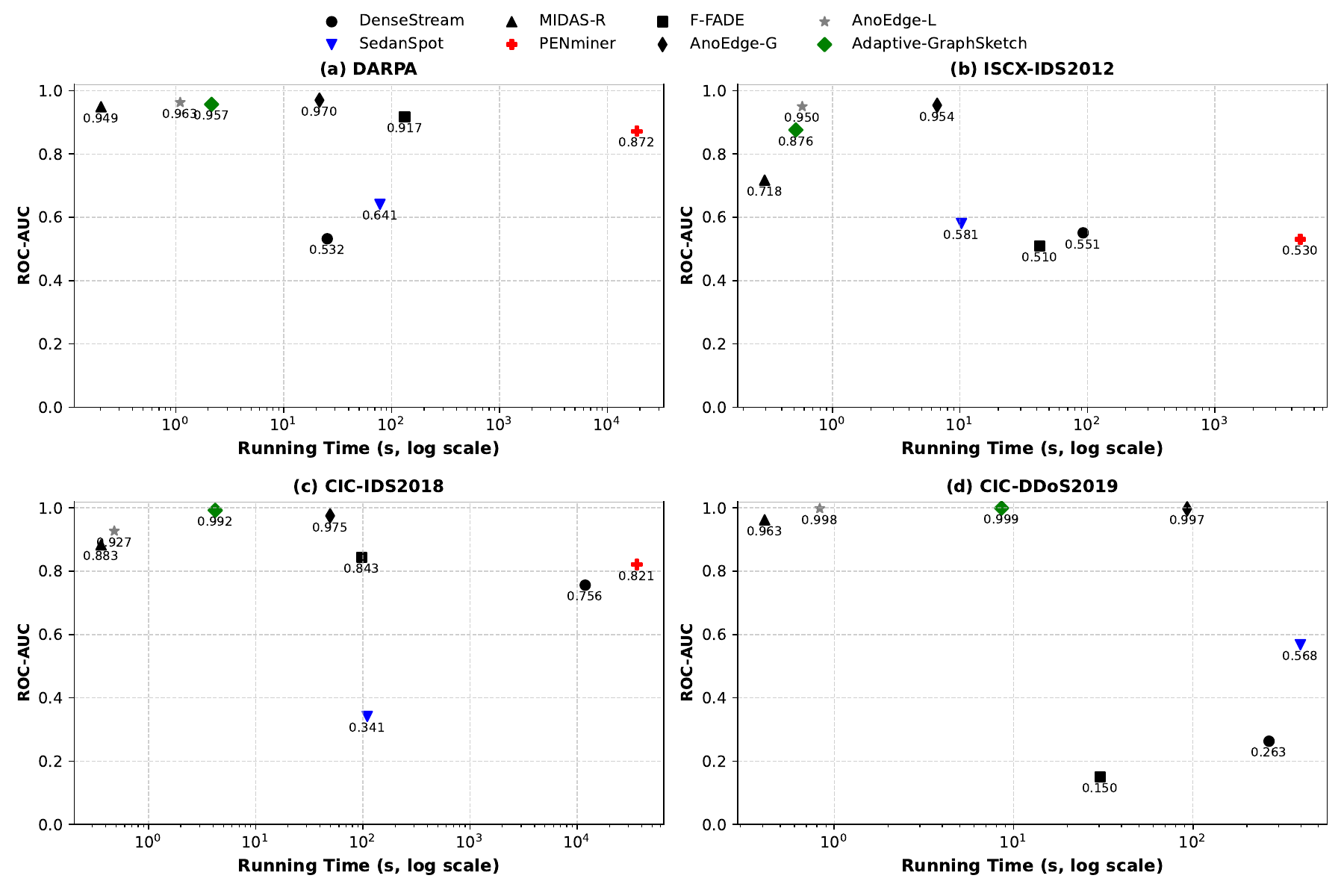}
    
    \caption{\textbf{AUC vs. Running Time (log scale) across four benchmark datasets. }\textsc{Adaptive-GraphSketch} achieves the best trade-off between accuracy and efficiency, outperforming state-of-the-art baselines across most settings.}
    \label{fig:auc_vs_runtime_all}
\end{figure*}


\subsection{Datasets}
\label{sec:datasets}
We experiment with four intrusion detection benchmarks, each serving as an ideal testbed for evaluating different aspects of streaming anomaly detection models.
DARPA \cite{d3_lippmann2000_DARPA} consists of $4.5M$ \textit{IP-IP} communications among $25.5K$ nodes over $46.5K$ discrete timestamps, offering a rich mix of attacks and temporal granularity. ISCX-IDS2012 \cite{shiravi2012toward_ISCX_IDS2012} includes $1.1M$ labeled flows over 165K timestamps, capturing stealthy infiltration and brute-force behaviors. CIC-IDS2018 \cite{sharafaldin2018toward_CIC_IDS2018} contains $7.9M$ edges among $33.1K$ nodes over $38.5K$ timestamps, covering a broad spectrum of modern attacks including botnets, DDoS, and port scans. CIC-DDoS2019 \cite{sharafaldin2019developing_CIC_IDS2019} contains $20.3M $edges, $1.29K$ unique nodes, and $12.2K$ timestamps, and it's characterized by high edge density and burst-heavy traffic. 
Table~\ref{tab:dataset_stats} summarizes the datasets with node, edge, and timestamp counts.


\subsection{Experimental Setup}
\label{sec:experimental_setup}

Experiments are conducted on a 13\textsuperscript{th} Gen Intel Core i9-13900 CPU (24 cores, 5.6 GHz), 32GB RAM, running Ubuntu 22.04. \textsc{Adaptive-GraphSketch}, implemented in C++, is compared against state-of-the-art baselines: \textsc{DenseStream} \cite{105_shin2017_DenseAlert}, \textsc{SedanSpot} \cite{101_eswaran2018_SedanSpot}, \textsc{MIDAS-R} \cite{040_bhatia2022_MIDAS_latest}, \textsc{PENminer} \cite{0103_belth2020_PENminer}, \textsc{F-FADE} \cite{043_F-FADE_chang2021f}, and \textsc{AnoEdge-G/L} \cite{056_bhatia2023_AnoEdge_sketch_based}, using their public implementations.

\textbf{Metric:} We report Area Under the ROC Curve (AUC) and runtime. AUC is calculated by plotting true positive rate (TPR) vs. false positive rate (FPR) at various thresholds and measuring the area under the curve. 
\textsc{Adaptive-GraphSketch} applies a temporal tensor sketch with decay and EWMA smoothing. 
Sketch parameters  includes: $r \in [2, 10]$ rows, $c \in [10, 1300]$ columns (e.g., $c=10$ for DARPA, $800$ for ISCX, $1300$ for CIC-DDoS2019). Decay $\gamma \in [0.95, 0.99]$, step size $\delta \in [10.0, 15.0]$, EWMA $\alpha \in [0.65, 0.95]$ depending on burst intensity, where lower $\alpha$ values are used for burst-heavy traffic (e.g., DDoS2019) to empahsize recent changes,  while higher $\alpha$ values (up to $0.95$) stabilizes scoring on  datasets with steadier flow e.g., DARPA. 

All experiments are repeated 5 times, and we report the average of  the ROC-AUC and  I/O runtimes in order to mitigate the effect of hash randomization.

\subsection{Accuracy}
\label{sec:accuracy}

Table~\ref{tab:auc_runtime_table} presents the ROC-AUC and runtime of \textsc{Adaptive-GraphSketch} compared to the established baselines across four benchmark datasets.


\textbf{Detection Performance:} \textsc{Adaptive-GraphSketch} demonstrates strong and consistent anomaly detection results across all four benchmarks, with AUC scores of 0.9568 (DARPA), 0.8761 (ISCX-IDS2012), 0.9923 (CIC-IDS2018), and 0.9993 (CIC-DDoS2019). While not the top performer on DARPA and ISCX-IDS2012, where \textsc{AnoEdge-G/L} achieve slightly higher AUC, our method offers superior overall efficiency and outperforms all baselines on real-time, large-scale datasets (CIC-IDS2018 and CIC-DDoS2019), with the advantage of balancing high accuracy and low runtime.

Compared to \textsc{MIDAS-R}, our model improves AUC by 1\% (DARPA), 22\% (ISCX), 12\% (CIC-IDS2018), and 4\% (CIC-DDoS2019). Against \textsc{F-FADE}, it outperforms by 4.3\%, 71.8\%, 17.7\%, and \textsc{566\%} respectively. Compared to \textsc{SedanSpot}, we observe significant performance of 49.3\% (DARPA), 51\% (ISCX), 190.7\% (CIC-IDS2018), and 76\% (CIC-DDoS2019). \textsc{PENminer}, while achieving reasonable AUC (e.g., 0.872 on DARPA, 0.821 on CIC-IDS2018), incurs extreme computational overhead (24+ hours on CIC-DDoS2019). Due to this limitation (reported in \cite{0103_belth2020_PENminer} and confirmed in our trials), we adopt its AUC values from the \textsc{AnoEdge} benchmark \cite{056_bhatia2023_AnoEdge_sketch_based}.

In contrast, our model outperforms \textsc{PENminer} in accuracy on all datasets (9.7\% on DARPA, 6.9\% on CIC-IDS2018, and 34\% on ISCX-IDS2012), and completes in less than 10 seconds. \textsc{AnoEdge-G} performs well (e.g., 0.970 on DARPA), but suffers from high latency (up to 92 seconds on CIC-DDoS2019), while \textsc{AnoEdge-L} is faster, it's less stable, with lower AUC (e.g., 0.927 on CIC-IDS2018). Overall, our model consistently achieves higher AUC on 3 out of 4 datasets, demonstrating both precision and robustness.

\textbf{Running Time:} Table~\ref{tab:auc_runtime_table} reports model runtimes. Our method is up to \textbf{36$\times$ faster} than \textsc{SedanSpot} and at least \textbf{5$\times$ faster} than \textsc{AnoEdge-G}, while maintaining comparable or higher accuracy. \textsc{F-FADE} shows severe training instability and degraded performance on large datasets, while \textsc{DenseStream} is slower and less scalable in dense networks. \textsc{PENminer} took over 24 hours on CIC-DDoS2019, making it impractical for real-time use. Overall, our model offers a better speed–accuracy trade-off, enabling real-time anomaly detection in high-velocity edge streams.

\begin{figure}[h]
    \centering

    \includegraphics[width=0.48\textwidth, height=0.22\textheight]{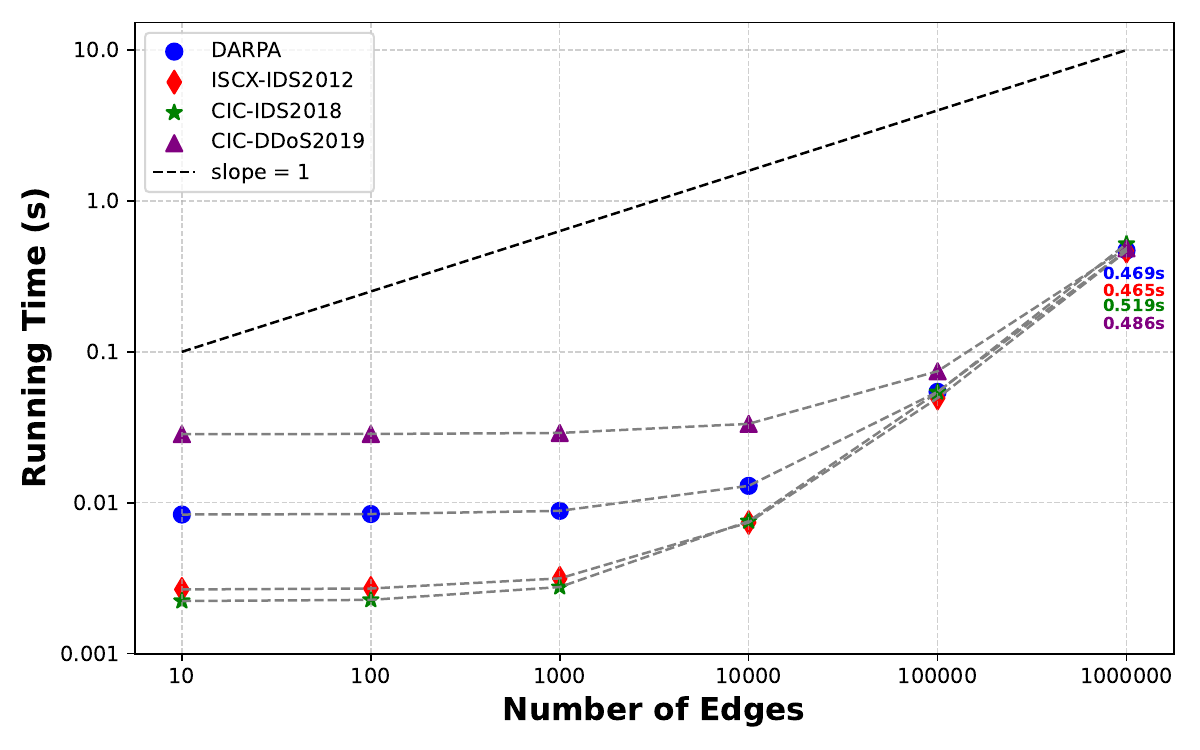}
    \caption{\textbf{Scalability of Adaptive-GraphSketch with Number of Edges}. Runtime increases with the number of edges from $10$ to $10^6$, and a slope-$1$ line is included for linearity comparison. Final values at $10^6$ edges are annotated below each marker.}
    \label{fig:graphsketch_scalability}
\end{figure}

\subsection{AUC vs. Running Time}
\label{sec:auc_vs_runtime}
To highlight the trade-off between accuracy and efficiency, 
Figure~\ref{fig:auc_vs_runtime_all} plots AUC against runtime (log scale, seconds) on four datasets. \textsc{Adaptive-GraphSketch} consistently achieves the highest AUC with much lower runtime. Compared to traditional baselines (e.g., MIDAS-R, SedanSpot, F-FADE), it is both faster and more accurate, and it outperforms or matches recent methods (e.g., AnoEdge-G, AnoEdge-L), showing a better balance of precision and scalability for real-time edge stream detection in large graphs.



\subsection{Scalability and Robustness}

We evaluate \textsc{Adaptive-GraphSketch} scalability on four datasets. 
Figure~\ref{fig:graphsketch_scalability} shows runtime (seconds) as edge volume grows from $10$ to $10^6$ on a log-log scale. 
A slope-$1$ reference line is included to benchmark linear growth.

\textsc{Adaptive-GraphSketch} scales near-linearly, processing 1M edges in under \textbf{0.52s} (\text{0.469s} DARPA, \text{0.464s} ISCX, \text{0.519s} CIC-IDS2018, \text{0.486s} CIC-DDoS2019), confirming our model's efficiency under both light and high-volume.



\begin{figure}[h]
    \centering
    \includegraphics[width=0.48\textwidth, height=0.22\textheight]{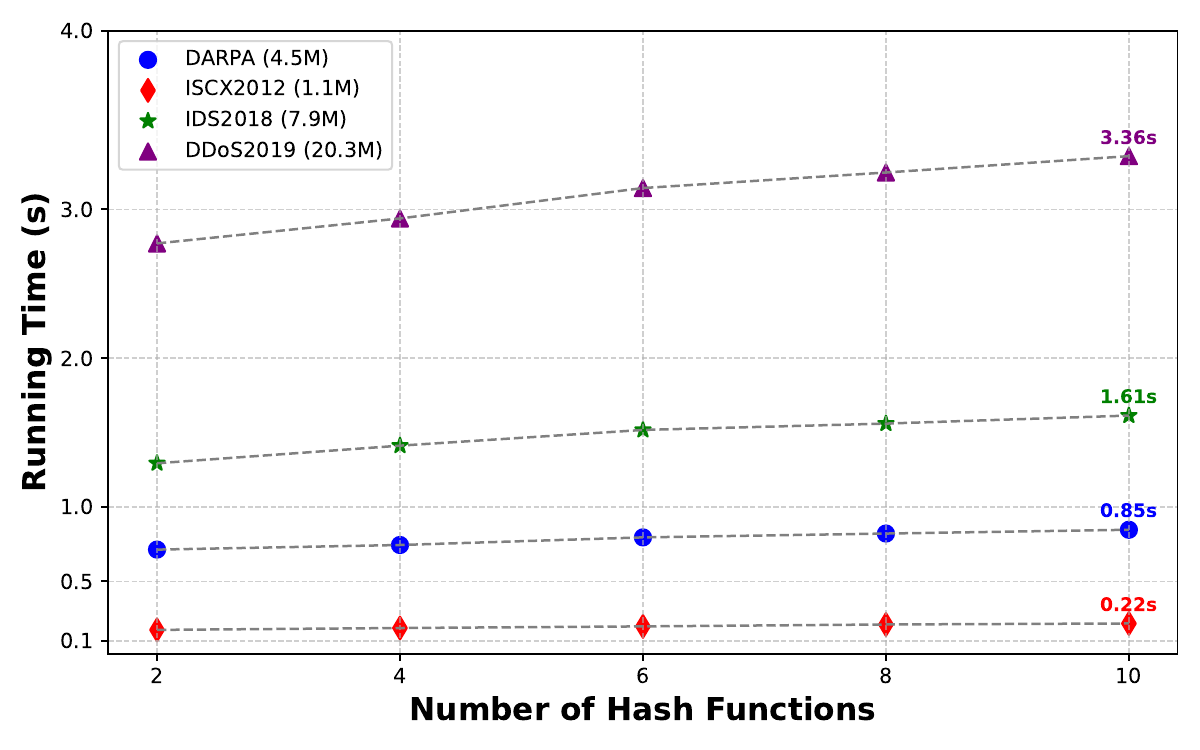}
    \caption{\textbf{Scalability of \textsc{Adaptive-GraphSketch} with Varying Hash Functions.} Runtime performance is evaluated on four datasets: DARPA (4.5M), ISCX2012 (1.1M), IDS2018 (7.9M), and DDoS2019 (20.3M). Each curve plots total processing time as the number of hash rows increases from 2 to 10. Final runtimes at $r = 10$ are annotated beside the respective markers.}
    \label{fig:hash_scalability}
\end{figure}
We further examine runtime sensitivity to the number of hash functions, a key parameter in sketch-based models. As shown in Figure~\ref{fig:hash_scalability}, increasing hash rows from 2 to 10 results in steady, linear runtime growth. Even at full dataset size (e.g., 20.3M edges for DDoS2019), processing remains efficient and completes in just \textbf{3.36s} with 10 hashes. This validates the robustness and scalability of our sketch design under increasing sketch complexity.

\subsection{Efficiency Analysis}

The efficiency of \textsc{Adaptive-GraphSketch} stems from its lightweight design that avoids costly operations. Unlike conventional methods such as random walks \cite{ekle2024_dynamicGraph2024}, and  matrix factorizations \cite{039c1_xie2023_multi_view}, our method performs constant-time operations per edge. Each edge is processed in \textit{constant time} via Count-Min Sketch with conservative updates \cite{estan2003new_CMS_Sketch_conservative_update}, and performs \textit{multi-sketch tensors}, where sketches are stored as compact time–protocol tensors to preserve temporal granularity while conserving memory. Furthermore, anomaly scores are smoothed using \textit{Bayesian exponential moving averages} for stability under dynamic patterns. Combined with CPU-level vectorization and pruning, these elements enable real-time detection with linear scalability, as confirmed by the results in Figures~\ref{fig:graphsketch_scalability} and \ref{fig:hash_scalability}.

\section{Conclusion}
\label{sec6:conclusion}

In this paper, we presented \textsc{Adaptive-GraphSketch}, a real-time anomaly detection framework that leverages tensor-based sketching and Bayesian smoothing for detecting edge-level anomalies in dynamic graphs. By integrating Count-Min Sketch with conservative updates, exponential decay, and EWMA-based aggregation, our method achieves robust detection with sub-second latency and low memory overhead.
Experiments across four real-world intrusion detection benchmark datasets  show that \textsc{Adaptive-GraphSketch} outperforms state-of-the-art streaming baselines such as \textsc{AnoEdge-G/L}, \textsc{MIDAS-R}, \textsc{PENminer} and \textsc{F-FADE}, achieving up to 6.5\% AUC gain on CIC-IDS2018 and up to 15.6\% on CIC-DDoS2019, while processing up to 20 million edges in under 3.4 seconds with 10 hashes. Unlike matrix factorization or random walk-based methods, our approach supports constant-time edge updates and scales linearly with edge volume and sketch complexity, making it suitable for edge computing and real-time applications. 

Future work includes extending the framework for multi-modal anomaly detection with content-aware sketches. We also aim to optimize runtime via parallelization and hardware acceleration, while addressing temporal drift and heterogeneity in evolving networks.


\section*{Acknowledgments}
We thank the CS Department at Tennessee Tech University and the Machine Intelligence and Data Science (MInDS) Center for providing resources to work on this project.


\bibliographystyle{IEEEtran}

\end{document}